\def\year{2017}
\documentclass[letterpaper]{article}
\usepackage{aaai}
\usepackage{times}
\usepackage{helvet}
\usepackage{courier}
\usepackage{amssymb}
\usepackage{booktabs}
\usepackage{diagbox}
\usepackage{amsmath, graphicx, algorithm, algpseudocode, amsmath,
  pdfpages, fancyhdr, paralist, booktabs, caption}
\usepackage{xcolor}
\usepackage{makecell}
\usepackage{colortbl}
\usepackage{tabularx}
\usepackage{paralist}
\usepackage{mdwlist}
\usepackage{multirow}
\usepackage{array}
\usepackage{subfig}
\usepackage{url}
\usepackage{tabularx}
\usepackage[colorinlistoftodos,color=orange!70,textsize=scriptsize,shadow]{todonotes}
\usepackage{pifont}

\newcolumntype{P}[1]{>{\centering\arraybackslash}p{#1}}

\setcellgapes{2pt}
\makegapedcells

\title{Run, Skeleton, Run: Skeletal Model in a Physics-Based Simulation}
\author{Mikhail Pavlov$^{\star}$
\quad Sergey Kolesnikov$^{\star}$
\quad Sergey M. Plis$^{\star}$ \\
$^{\star}$Reason8 
}

\graphicspath{
{figures/}
}

\newcommand{\A}{\mathcal{A}}
\newcommand{\R}{\mathbb{R}}
\newcommand{\E}{\mathbb{E}}

\renewcommand{\L}[1]{L^{CLIP}(#1)}

\definecolor{tableShade2}{HTML}{ECF3FE} %Finder

\begin{document}
\maketitle

\begin{abstract}
\begin{quote}
In this paper, we present an approach to solve a physics-based reinforcement learning challenge ``Learning to Run'' with objective to train physiologically-based human model to navigate a complex obstacle course as quickly as possible.
The environment is computationally expensive, has a high-dimensional continuous action space and is stochastic.
We benchmark state of the art policy-gradient methods and test several improvements, such as layer normalization, parameter noise, action and state reflecting, to stabilize training and improve its sample-efficiency.
We found that the Deep Deterministic Policy Gradient method is the most efficient method for this environment and the improvements we have introduced help to stabilize training.
Learned models are able to generalize to new physical scenarios, e.g. different obstacle courses.
\end{quote}
\end{abstract}

\noindent 

\section{Introduction}
Reinforcement Learning (RL)~\cite{sutton1998reinforcement} is a significant subfield of Machine Learning and Artificial Intelligence along with the supervised and unsupervised subfields with numerous applications ranging from trading to robotics and medicine.
It has already achieved high levels of performance on Atari games~\cite{mnih2015human}, board games~\cite{silver2016mastering} and 3D navigation tasks~\cite{mnih2016asynchronous,jaderberg2016reinforcement}.

All of above tasks have one feature in common - there is always some well-defined reward function, for example, game score, which can be optimized to produce the required behaviour.
Nevertheless, there are are many other tasks and environments, for which it is still unclear what is the ``correct'' reward function to optimize.
And it is even a harder problem, when we talk about continuous control tasks, such as physics-based environments~\cite{todorov2012mujoco} and robotics~\cite{gu2017deep}.

Yet, recently a substantial interest is directed to research employing
physics-based  based environment. 
These environments are significantly more interesting, challenging and
realistic than the well defined games; at the same time they are still
simpler than real conditions with physical agents, while being cheap
and more accessible.
One of the interesting researches is the work
of~\citeauthor{schulman2015high} where a simulated robot learned to
run and get up off the ground~\cite{schulman2015high}.
Another paper is by \citeauthor{heess2017emergence} where the authors trained several simulated bodies on a diverse set of challenging terrains and obstacles, using a simple reward function based on forward progress~\cite{heess2017emergence}.

To solve the problem of continuous control in simulation environments it has become generally accepted to adapt the reward signal for specific environment.
Still it can lead to unexpected results when the reward function is modified even slightly, and for more advanced behaviors the appropriate reward function is often non-obvious.
To address this problem, the community came up with several environment-independent approaches such as unsupervised auxiliary tasks~\cite{jaderberg2016reinforcement} and unsupervised exploration rewards~\cite{pathak2017curiosity}.
All these suggestions are trying to solve the main challenge of reinforcement learning: how an agent can learn for itself, directly from a limited reward signal, to achieve best performance.

Besides the difficulty in defining the reward function, physically realistic environments usually have a lot of stochasticity, are computationally very expensive, and have high-dimensional action spaces.
To support learning in such settings it is necessary to have a reliable, scalable and sample-efficient reinforcement learning algorithm.
In this paper we evaluate several existing approaches and then improve upon the best performing approach for a physical simulator setting.
We present the approach that we have used to solve the ``Learning to
run'' -- NIPS 2017 competition
challenge\footnote{\url{https://www.crowdai.org/challenges/nips-2017-learning-to-run}}
with an objective to learn to control a physiologically-based human
model and make it run as quickly as possible.
The model that we present here has won the third place at the
challenge: \url{https://www.crowdai.org/challenges/nips-2017-learning-to-run/leaderboards}.

This paper proceeds as follows: first we review the basics of reinforcement learning, then we   describe environment used in challenge and models used in our experiment, after that we present results of our experiments and finally we discuss the results and conclude the work.

\section{Background}
We approach the problem in a basic RL setup of an agent interacting with an environment.
The ``Learning to run'' environment is fully observable and thus can be modeled as a Markov Decision Process (MDP)~\cite{bellman1957markovian}.
MDP is defined as a set of states (${\cal S} : \{s_i\}$), a set of actions ($\A : \{a_i\}$), a distribution over initial states $p(s_0)$, a reward function $r : {\cal S} \times \A \rightarrow \R$, transition probabilities $p(s_{t+1}|s_t,a_t)$, time horizon $T$, and a discount factor $\gamma \in [0, 1)$.
  A policy parametrized by $\theta$ is denoted with $\pi_\theta$.
  The policy can be either deterministic, or stochastic.
  The agent's goal is to maximize the expected discounted return $\eta(\pi_\theta) = \E_{\tau}[\sum_{t=0}^T \gamma_t r(s_t, a_t)]$, where $\tau = (s_0, a_0, \dots ,s_{T} )$ denotes a trajectory with $s_0 \sim p(s_0)$, $a_t \sim \pi_\theta(a_t|s_t)$, and $s_t \sim p(s_{t}|s_{t-1},a_{t-1})$.

\section{Environment}
\label{sec:env}
\begin{figure}
  \includegraphics[width=\columnwidth]{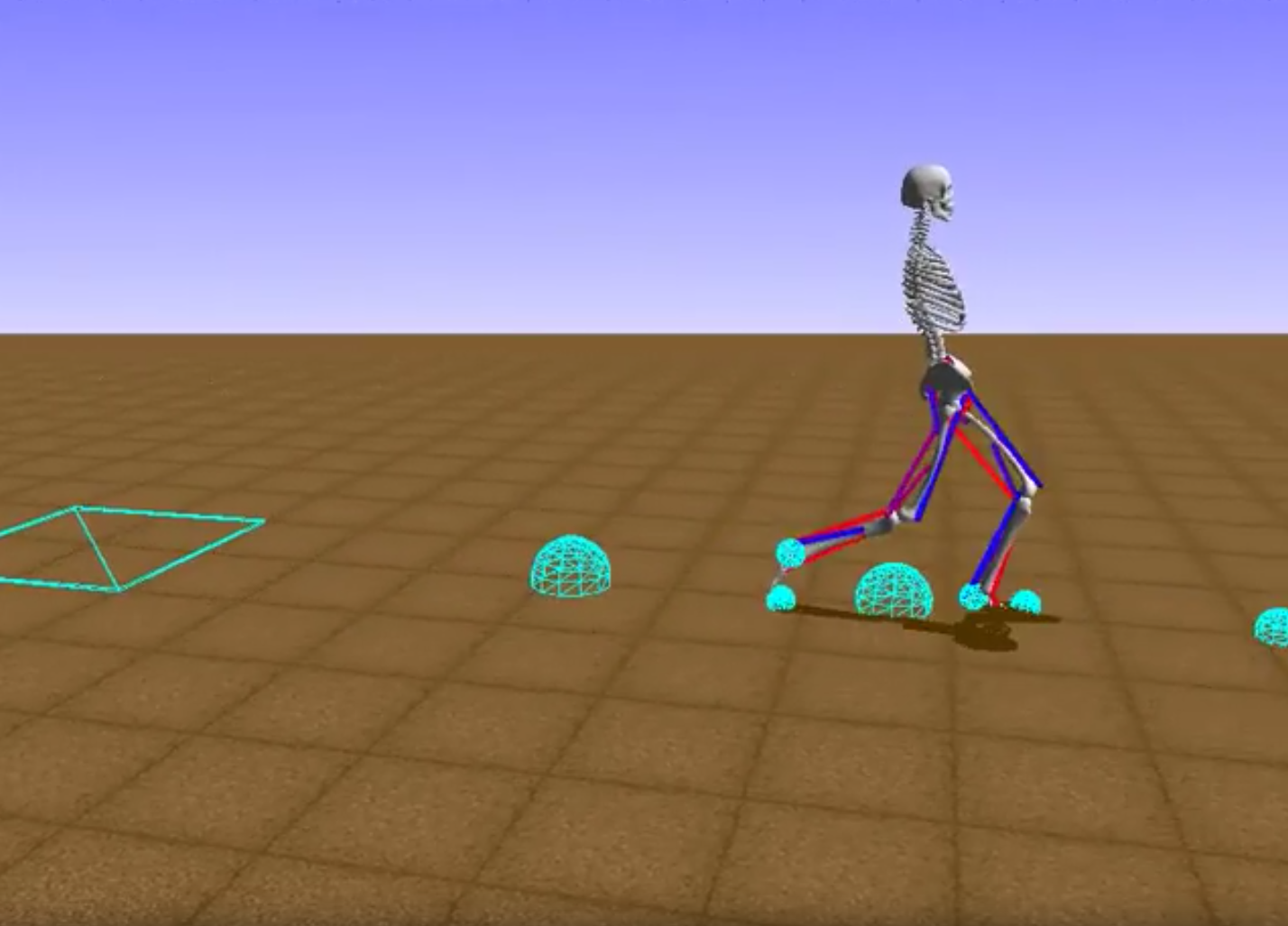}  
\caption{OpenSim screenshot that demonstrates the agent.}
\label{fig:screenshot}
\end{figure}
The environment is a musculoskeletal model that includes body segments
for each leg, a pelvis segment, and a single segment to represent the
upper half of the body (trunk, head, arms).
See Figure~\ref{fig:screenshot} for a clarifying screenshot.
The segments are connected with joints (e.g., knee and hip) and the motion of these joints is controlled by the excitation of muscles.
The muscles in the model have complex paths (e.g., muscles can cross more than one joint and there are redundant muscles).
The muscle actuators themselves are also highly nonlinear.

The purpose is to navigate a complex obstacle course as quickly as
possible. The agent operates in a 2D world. The obstacles are balls
randomly located along the agent's way. Simulation is done using
OpenSim~\cite{delp2007opensim} library which relies on the Simbody~\cite{sherman2011simbody} physics engine.
The environment is described in Table~\ref{tab:opensim}. More detailed description of
environment  can be found on competition github page.\footnote{\url{https://github.com/stanfordnmbl/osim-rl}}

Due to a complex physics engine the environment is quite slow compared
to standard locomotion environments~\cite{todorov2012mujoco,roboschool:site}.
Some steps in environment could take seconds. Yet, the other
environments can be as fast as three orders of magnitudes faster.\footnote{\url{https://github.com/stanfordnmbl/osim-rl/issues/78}}
So it is crucial to train agent using the most sample-efficient method.

\begin{table}
\caption{Description of the OpenSim environment.}
\begin{tabularx}{\columnwidth}{r|X}
  \toprule
  parameters & description \\
  \midrule
  state $(s_t)$ & $\R^{41}$, coordinates and velocities of various body
                  parts and obstacle locations. All $(x, y)$
                  coordinates are absolute. To  improve
                  generalization of our controller and use data more
                  efficiently, we modified the original version of
                  environment making all $x$ coordinates relative to
                  the $x$ coordinate of pelvis. \\
  action $(a_t)$ & $\R^{18}$, muscles activations, 9 per leg, each in
                   $[0, 1]$ range. \\
  reward & $\R$, change in $x$ coordinate of pelvis plus  a small penalty for using ligament forces.\\
  terminal state & agent falls (pelvis $x < 0.65$) or 1000 steps in environment\\
  stochasticity & \begin{itemize*} 
  \item random strength of the psoas muscles
  \item random location and size of obstacles
  \end{itemize*}\\
  \bottomrule
\end{tabularx}
\label{tab:opensim}
\vspace{-0.5cm}
\end{table}

\section{Methods}
In this section we briefly describe the models we have evaluated in the task of the ``Learning to run'' challenge.
We also describe our improvements to the model best performing in the competition: Deep Deterministic Policy Gradient (DDPG)~\cite{lillicrap2015continuous}.

\subsection{On-policy methods}
On-policy RL methods can only update agent's behavior with data generated by the current policy.
We consider two popular on-policy algorithms, namely Trust Region Policy Optimization (TRPO)~\cite{schulman2015trust} and Proximal Policy Optimization (PPO)~\cite{schulman2017proximal} as the baseline algorithms for environment solving.

\subsubsection{Trust Region Policy Optimization}
(TRPO) is one of the notable state-of-the-art RL algorithms, developed by~\citeauthor{schulman2015trust}, that has theoretical monotonic improvement guarantee. As a basis, TRPO~\cite{schulman2015trust} using REINFORCE~\cite{williams1992simple} algorithm, that estimates the gradient of expected return $\nabla_{\theta}\eta(\pi_{\theta})$ via likelihood ratio:
\begin{align}
  \nabla_{\theta}\eta(\pi_{\theta}) &=
                                      \frac{1}{NT}\sum_{i=1}^N\sum_{t=0}^T
                                      \nabla_\theta \log\pi_\theta(a_t^i|s_t^i)(R_t^i-b_t^i),
\end{align}
where $N$ is the number of episodes, $T$ is the number of steps per
episode, $R_t^i = \sum_{t^{'}=t}^{T}\gamma^{t^{'}-t}r_{t^{'}}^i$ is
the cumulative reward and $b_t^t$ is a variance reducing
baseline~\cite{duan2016benchmarking}.
After that, an ascent step is taken along the estimated gradient. 
TRPO improves upon REINFORCE by computing an ascent direction that
ensures a small change in the policy distribution.
As the baseline TRPO we have used the agent described in~\cite{schulman2015trust}.

\subsubsection{Proximal Policy Optimization}
(PPO) as TRPO tries to estimate an ascent direction of gradient of expected return that restricts the changes in policy to small values.
We used clipped surrogate objective variant of proximal policy optimization~\cite{schulman2017proximal}.
This modification of PPO is trying to compute an update at each step
that minimizes following cost function:
\begin{align}
\L{\theta}  & = \hat{\E}_t[\min(r_t(\theta)\hat{A}_t,\text{clip}(r_t(\theta),1-\epsilon,1+\epsilon)\hat{A}_t)],
\end{align}
where $r_t(\theta) =
\frac{\pi_\theta(a_t|s_t)}{\pi_{\theta^{old}}(a_t|s_t)}$ is a
probability ratio  (the new divided by the old policy), $\hat{A}_t = R_t-b_t$ is
empirical return minus the baseline. 
This cost function is very easy to implement and  allows multiple epochs of minibatch updates.

\subsection{Off-policy methods}
In contrast to on-policy algorithms, off-policy methods allow learning based on all data from arbitrary policies.
It significantly increases sample-efficiency of such algorithms relative to on-policy based methods.
Due to simulation speed litimations of the environment, we will only consider  Deep Deterministic Policy Gradient (DDPG)~\cite{lillicrap2015continuous}.

\subsubsection{Deep Deterministic Policy Gradient}
(DDPG) consists of actor and critic networks. 
Critic is trained using Bellman equation and off-policy data:
\begin{align}
  Q(s_t,a_t) &= r(s_t,a_t) + \gamma Q(s_{t+1}, \pi_\theta(s_{t+1})),
\end{align}
where $\pi_\theta$ is the actor policy.
The actor is trained to maximize the critic's estimated Q-values by back-propagating through critic and actor networks.
As in original article we used replay buffer and the target network to stabilize training and more efficiently use samples from environment.

\subsubsection{DDPG improvements}
Here we present our improvements to the DDPG method.
We used some standard reinforcement learning techniques: action repeat (the agent selects action every 5th state and selected action is repeated on skipped steps) and reward scaling. After several attempts, we choose a scale factor of 10 (i.e. multiply reward by ten) for our experiments.
For exploration we used Ornstein-Uhlenbeck (OU) process~\cite{uhlenbeck1930theory} to generate temporally correlated noise for efficient exploration in physical environments.
Our DDPG implementation was parallelized as follows: $n$ processes collected samples with fixed weights all of which were processed by the learning process at the end of an episode, which updated their weights.
Since DDPG is an off-policy method, the stale weights of the samples only improved the performance providing each sampling process with its own weights and thus improving exploration.

\subsubsection{Parameter noise}
Another improvement is the recently proposed parameters
noise~\cite{plappert2017parameter} that perturbs network weights
encouraging state dependent exploration.
We used parameter noise only for the actor network.
Standard deviation $\sigma$ for the Gaussian noise is chosen according
to the original work~\cite{plappert2017parameter} so that measure $d$:
\begin{align}
  d(\pi,\widetilde{\pi}) = \sqrt{\bigg(\frac{1}{N}\sum_{i=1}^N\E_s[(\pi(s)_i - \widetilde{\pi}(s)_i)^2]\bigg)},
\end{align}
where $\widetilde{\pi}$ is the policy with noise, equals to $\sigma$ in OU.
For each training episode we switched between the action noise and the parameter noise choosing them with 0.7 and 0.3 probability respectively.

\subsubsection{Layer norm}
\citeauthor{henderson2017deep} showed that layer normalization~\cite{ba2016layer} stabilizes the learning process in a wide range of reward scaling.
We have investigated this claim in our settings.
Additionally, layer normalization allowed us to use same perturbation scale across all layers despite the use of parameters noise~\cite{plappert2017parameter}.
We normalized the output of each layer except the last for critic and actor by standardizing the activations of each sample.
We also give each neuron its own adaptive bias and gain. 
We applied layer normalization before the nonlinearity.

\subsubsection{Actions and states reflection symmetry}
The model has  bilateral body symmetry. State components and actions
can be reflected to increase sample size by factor of 2.  We sampled
transitions from replay memory, reflected states and actions and used
original states and actions as well as reflected  as batch in training
step. This procedure improves stability of learned policy. If we don’t
use this step our model learned suboptimal policies, when for example
muscles for only one leg are active and other leg just follows first
leg. 

\section{Results}

It this section we presents our experiments and setup.
For all experiments we used environment with 3 obstacles and random strengths of the psoas muscles.
We tested models on setups running 8 and 20 threads.
For comparing different PPO, TRPO and DDPG settings we used 20 threads per model configuration.
We have compared various combinations of improvements of DDPG in two identical settings that only differed in the number of threads used per configuration: 8 and 20.
The goal was to determine whether the model rankings are consistent when the number of threads changes.
For $n$ threads (where $n$ is either 8 or 20) we used $n-2$ threads for sampling transitions, 1 thread for training, and 1 thread for testing.
For all models we used identical architecture of actor and critic networks.
All hyperparameters are listed in Table~\ref{tab:hyperopt}.
Our code used for competition and described experiments can be found
in a github repo.\footnote{Theano:
  \url{https://github.com/fgvbrt/nips_rl} and  PyTorch: \url{https://github.com/Scitator/Run-Skeleton-Run}} 
Experimental evaluation is based on the undiscounted return $\E_{\tau} [ \sum_{t=0}^T r(s_t, a_t)]$.

\begin{table}
\caption{Hyperparameters used in the experiments.}
\begin{tabularx}{\columnwidth}{r|X}
  \toprule
  parameters & Value \\
  \midrule
  Actor network architecture &  $[64,64]$, elu activation\\
  Critic network architecture & $[64,32]$, tanh activation \\
  Actor learning rate & linear decay from $1\mathrm{e}{-3}$ to
                        $5\mathrm{e}{-5}$ in $10\mathrm{e}{6}$ steps with Adam optimizer\\
  Critic learning rate & linear decay from $2\mathrm{e}{-3}$ to
                        $5\mathrm{e}{-5}$ in $10\mathrm{e}{6}$ steps with Adam optimizer\\
  Batch size & 200\\
  $\gamma$ & 0.9 \\
  replay buffer size & $5\mathrm{e}{6}$\\
  rewards scaling & 10 \\
  parameter noise probability & 0.3 \\
  OU exploration parameters & $\theta = 0.1$, $\mu = 0$, $\sigma = 0.2$,
                          $\sigma_{min}=0.05$, $dt = 1\mathrm{e}{-2}$,
                          $n_\text{steps}$ annealing
                          $\sigma_\text{decay} 1\mathrm{e}{6}$ per thread\\
                          \bottomrule
\end{tabularx}
\label{tab:hyperopt}
\end{table}

\subsection{Benchmarking different models}
\begin{figure}[ht]
\includegraphics[width=\columnwidth]{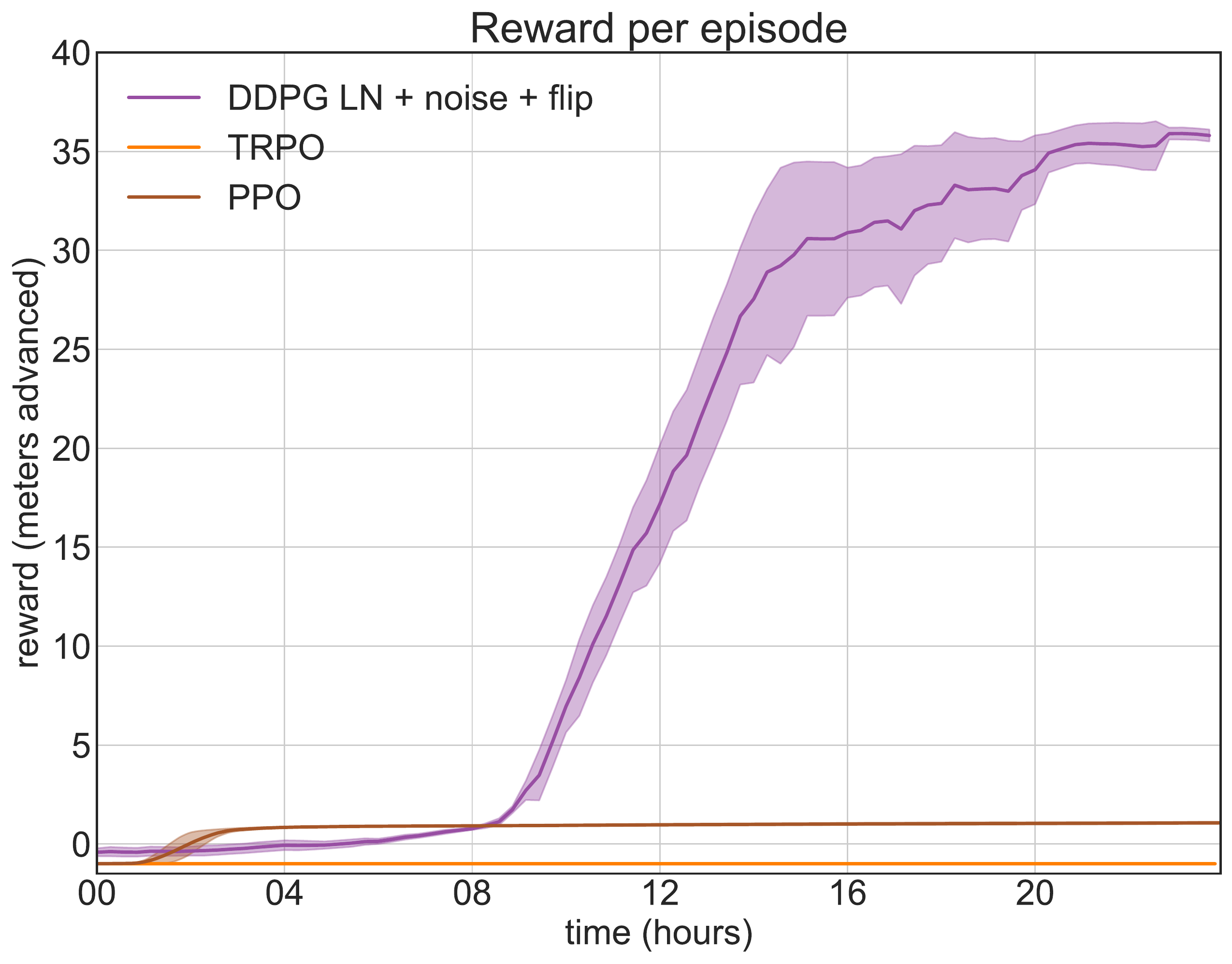}
\caption{Comparing test reward of the baseline models and the best performing model that we have used in the ``Learning to run'' competition.}
\label{fig:baseline_compare}
\end{figure}

Comparison of our winning model with the baseline approaches is presented in Figure~\ref{fig:baseline_compare}.
Among all methods the DDPG significantly outperformed PPO and TRPO.
The environment is time expensive and method should utilized experience as effectively as possible.
DDPG due to experience replay (re)uses each sample from environment many times making it the most effective method for this environment.

\subsection{Testing improvements of DDPG}

\begin{figure*}[ht!]%
        \centering
        \subfloat[$8$ threads]{
                \includegraphics[width=0.475\textwidth]{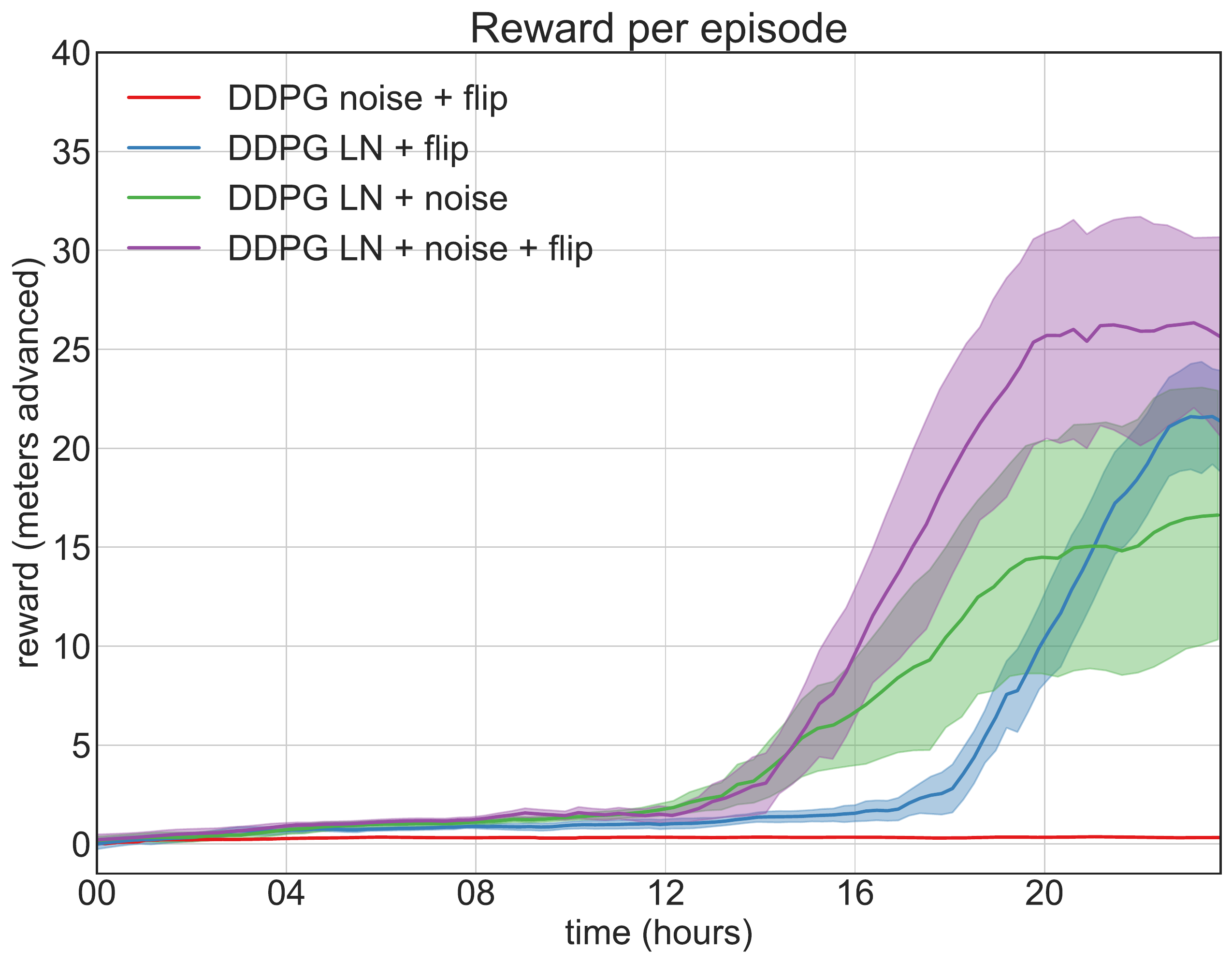}
                \label{fig:8_threads}
        }
        \quad
        \subfloat[$20$ threads]{
                \includegraphics[width=0.475\textwidth]{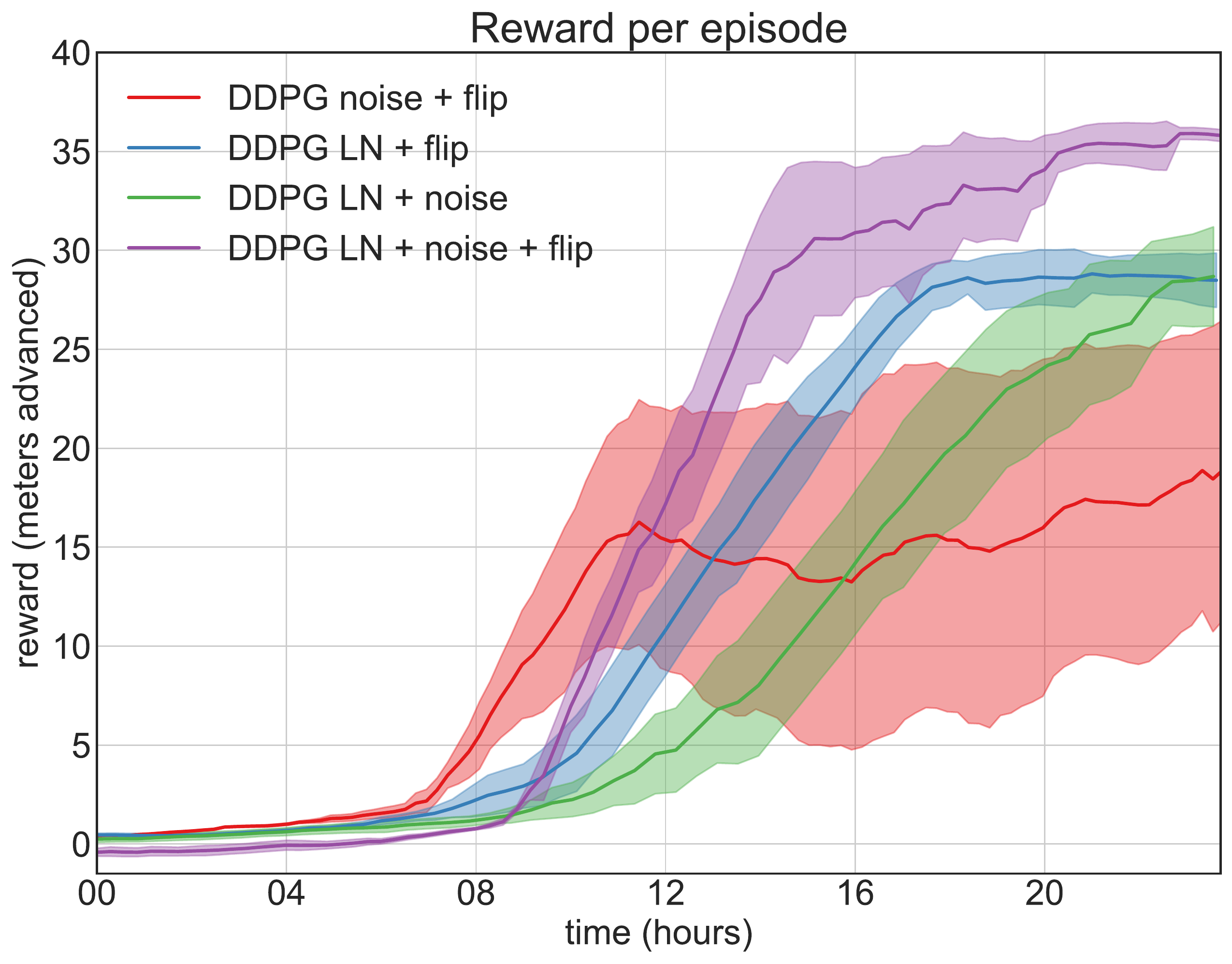}
                \label{fig:20_threads}
        }
        \caption{Comparing test reward for various modifications of the DDPG algorithm with 8 threads per configuration (Figure~\ref{fig:8_threads}) and 20 threads per configuration (Figure~\ref{fig:20_threads}). Although the number of threads significantly affects performance, the model ranking approximately stays the same.}
        \label{fig:ablation}
\end{figure*}

To evaluate each component we used an ablation study as it was done in the rainbow article~\cite{hessel2017rainbow}.
In each ablation, we removed one component from the full combination.
Results of experiments are presented in Figure~\ref{fig:8_threads} and Figure~\ref{fig:20_threads} for 8 and 20 threads respectively.
The figures demonstrate that each modification leads to a statistically significant performance increase.
The model containing all modifications scores the highest reward.
Note, the substantially lower reward in the case, when parameter noise was employed without the layer norm.
One of the reasons is our use of the same perturbation scale across all layers, which does not work that well without normalization.
Also note, the behavior is quite stable across number of threads, as
well as the model ranking.
As expected, increasing the number of threads improves the result.

Maximal rewards achieved in the given time for 8 and 20 threads cases for each of the combinations of the modifications is summarized in Table~\ref{tab:maxreward}.
The main things to observe is a substantial improvement effect of the number of threads, and stability in the best and worst model rankings, although the models in the middle are ready to trade places.

\begin{table}
\caption{Best achieved reward for each DDPG modification.}
\begin{tabularx}{\columnwidth}{c|X|X}
  \toprule
  \backslashbox{agent}{\# threads} & 8 & 20 \\
  \midrule
  DDPG + noise + flip & 0.39 & 23.58 \\
  DDPG + LN + flip & 25.29 & 31.91\\
  DDPG + LN + noise & 25.57 & 30.90 \\
  DDPG + LN + noise + flip & \bf{31.25} & \bf{38.46}\\
  \bottomrule
\end{tabularx}
\label{tab:maxreward}
\end{table}

%\vspace{-1cm}
\section{Conclusions}
%\vspace{-1cm}
Our results in OpenSim experiments indicate that in a computationally expensive stochastic environments that have high-dimensional continuous action space the best performing method is off-policy DDPG.
We have tested 3 modifications to DDPG and each turned out to be important for learning.
Action states reflection doubles the size of the training data and improves stability of learning and encourages the agent to learn to use left and right muscles equally well.
With this approach the agent truly learns to run.
Examples of the learned policies with and without the reflection are present at this URL \url{https://tinyurl.com/ycvfq8cv}.
Parameter and Layer noise additionally improves stability of learning due to introduction of state dependent exploration.
In general, we believe that investigation of human-based agents in physically realistic environments is a promising direction for future research.

%\section{Discussion}
%\section{Acknowledgments}

\pagebreak
%\section{References}

\bibliographystyle{aaai}
\bibliography{papers}

\end{document}